\documentclass[10pt,twocolumn,letterpaper]{article}

\usepackage[pagenumbers]{cvpr} 

\usepackage{graphicx}
\usepackage{amsmath}
\usepackage{amssymb}
\usepackage{booktabs}

\PassOptionsToPackage{numbers,sort,compress}{natbib}
\usepackage[utf8]{inputenc} 
\usepackage[T1]{fontenc}    
\usepackage{amsfonts}       
\usepackage{nicefrac}       
\usepackage{microtype}      
\usepackage{times}
\usepackage{epsfig}
\usepackage{bm}
\usepackage[pagebackref,breaklinks,colorlinks]{hyperref}
\usepackage{latexsym}
\usepackage{url}
\usepackage{soul}
\usepackage{multirow}
\usepackage[dvipsnames,table]{xcolor}
\usepackage{enumerate}
\usepackage{xparse}
\usepackage{float}
\usepackage{etoolbox}
\usepackage{mathtools}

\newcommand{\ctab}[1]{\begin{tabular}{c} #1 \end{tabular}}

\usepackage{makecell}

\definecolor{themered}{HTML}{FF8375}
\definecolor{green}{HTML}{22a6b3}
\definecolor{blue}{HTML}{3867d6}
\definecolor{yellow}{HTML}{f0932b}

\NewDocumentCommand{\Narendra}{ mO{} }{\textcolor{themered}{\textsuperscript{\textit{Narendra}}\textsf{\textbf{\small[#1]}}}}
\NewDocumentCommand{\Xiaodan}{ mO{} }{\textcolor{green}{\textsuperscript{\textit{Xiaodan}}\textsf{\textbf{\small[#1]}}}}
\NewDocumentCommand{\Chuhang}{ mO{} }{\textcolor{blue}{\textsuperscript{\textit{Chuhang}}\textsf{\textbf{\small[#1]}}}}
\NewDocumentCommand{\Jaechul}{ mO{} }{\textcolor{yellow}{\textsuperscript{\textit{Jaechul}}\textsf{\textbf{\small[#1]}}}}

\usepackage[capitalize]{cleveref}
\crefname{section}{Sec.}{Secs.}
\Crefname{section}{Section}{Sections}
\Crefname{table}{Table}{Tables}
\crefname{table}{Tab.}{Tabs.}

\def\cvprPaperID{8655} 
\def\confName{CVPR}
\def\confYear{2023}

\usepackage[toc,page]{appendix}

\begin{document}

\title{Language-driven Description Generation and Common Sense Reasoning for Video Action Recognition}

\author{Xiaodan Hu\textsuperscript{\textnormal{1}}, Chuhang Zou\textsuperscript{\textnormal{2}}, Suchen Wang\textsuperscript{\textnormal{2}}, Jaechul Kim\textsuperscript{\textnormal{2}}, Narendra Ahuja\textsuperscript{\textnormal{1}}\\
\textsuperscript{1}University of Illinois at Urbana-Champaign  \textsuperscript{2}Amazon.com LLC\\
\texttt{\fontfamily{pcr}\selectfont\{xiaodan8,n-ahuja\}@illinois.edu},  \texttt{\fontfamily{pcr}\selectfont\{zouchuha,jaechulk\}@amazon.com}
}
\maketitle

\begin{abstract}

Recent video action recognition methods have shown excellent performance by adapting large-scale pre-trained language-image models to the video domain. However, language models contain rich common sense priors — the scene contexts that humans use to constitute an understanding of objects, human-object interactions, and activities — that have not been fully exploited. In this paper, we introduce a framework incorporating language-driven common sense priors to identify cluttered video action sequences from monocular views that are often heavily occluded.
We propose: (1) A video context summary component that generates candidate objects, activities, and the interactions between objects and activities; (2) A description generation module that describes the current scene given the context and infers subsequent activities, through auxiliary prompts and common sense reasoning; (3) A multi-modal activity recognition head that combines visual and textual cues to recognize video actions. We demonstrate the effectiveness of our approach on the challenging Action Genome and Charades datasets. 


\end{abstract}

\section{Introduction}
\label{sec:introduction}

\begin{figure}[!ht]
\centering
\begin{tabular}{c}
\includegraphics[width=0.475\textwidth,keepaspectratio]{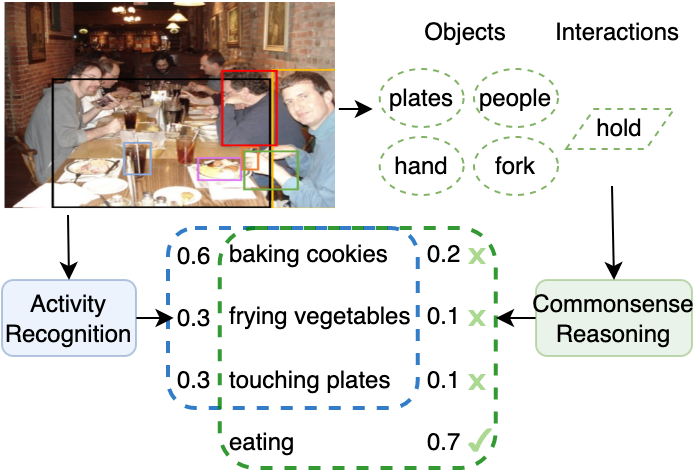}
\end{tabular}
\vspace{-10pt}
\caption{
Illustration of how \textbf{Common Sense Reasoning} aids video understanding. Vision systems may fail to recognize actions from cluttered and heavily occluded scenes. We propose to leverage common sense reasoning by summarizing and describing the scene context of objects, interactions, and activities, which can help suppress illogical predictions.
}
\vspace{-10pt}
\label{fig:example}
\end{figure}

With one glance at a video clip, we can effortlessly tell  the triggering event of the clip and the story that follows. This is because humans can make common sense reasoning to construct their understanding of the activities and objects presented. For example, if we see a person sitting at a table with a plate and food on it, our common sense automatically infers that the person intends to eat. While this task is easy for humans, it is not so trivial for today's vision system (as shown in Figure~\ref{fig:example}), especially for scenes with cluttered and temporally overlapping action sequences, as well as heavily occluded monocular views (e.g. Action Genome dataset~\cite{AG}).

\begin{figure*}[!ht]
\centering
\begin{tabular}{c}
\includegraphics[width=\textwidth,keepaspectratio]{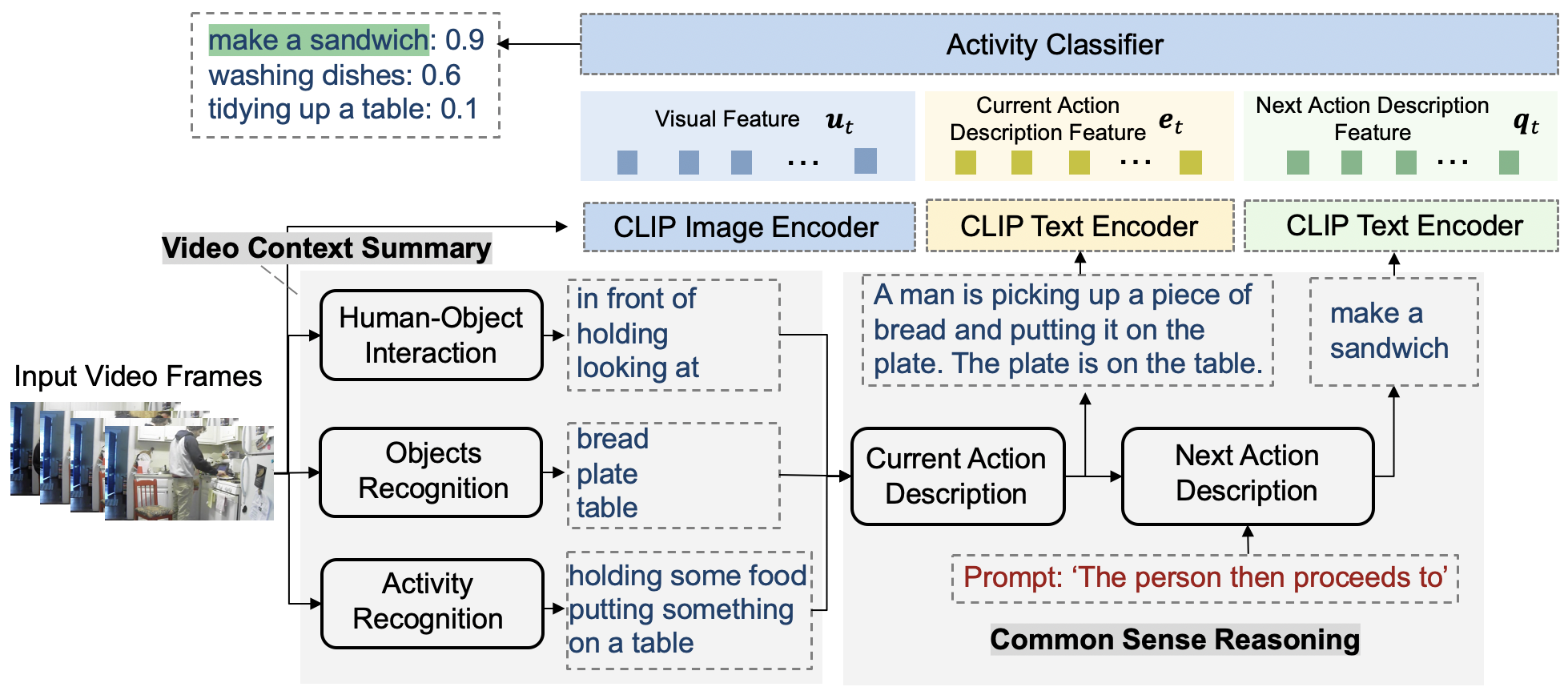}
\end{tabular}
\vspace{-10pt}
\caption{
\textbf{Overview} of proposed framework. The model contains three main components to understand the video: Video Context Summary $\mathcal{S}=\{\mathcal{S}_{int}, \mathcal{S}_{obj}, \mathcal{S}_{act}\}$, Common Sense Reasoning $\mathcal{C}=\{\mathcal{C}_{D}, \mathcal{C}_{C}\}$ and Multi-modal Activity Recognition $\mathcal{F}$. Given a sequence of video frames $\{I_t\}^{T-1}_{t=0}$, the Context Summary component $\mathcal{S}$ extracts the entire video activities $\hat{v}$ and per-frame human-object interactions $\{\hat{r}_t\}^{T-1}_{t=0}$ and objects $\{\hat{o}_t\}^{T-1}_{t=0}$. 
Then, in the Common Sense Reasoning component, the extracted context $\hat{r}_t, \hat{o}_t$ and $\hat{v}$ accompanying prompt templates are used as the input of the current action description generator $\mathcal{C}_{D}$, which generates the situation description of each frame $s^d_t$.
Given the description $s^d_t$, the subsequent action prediction module predicts the next possible activity $s^c_t$ with the help of additional prompt input. Finally, the Multi-modal Activity Recognition component $\mathcal{F}$ extracts image embeddings, current and subsequent descriptions embeddings, $(\bf{u}, \bf{e}, \bf{q})$, and concatenates them as input to the activity classifier to predict activity for each frame $\tilde{\bf{y}}_t$. 
We aggregate the predicted activity of each frame $\tilde{\bf{y}}_t$ to obtain our final predictions of the whole video $\hat{\bf{y}}$.
}
\vspace{-10pt}
\label{fig:overview}
\end{figure*}


Inspired by recent progress in learning large-scale language models~\cite{OPT,RoBERTa,chatgpt4o,google2024gemini2flash,bai2025qwen25vl} for near-human-level writing and reasoning abilities, we propose an activity recognition framework that marries visual cues such as object and interaction detection with linguistic descriptions that directly carry human common sense knowledge of the observed scene context. Our language-driven approach differs from recent video action recognition methods~\cite{ju2022prompting,Qu_2024_CVPR, Maaz2023VideoChatGPT, Qu_2025_CVPR, azad2025hierarq}, which directly apply contrastive pre-trained Multimodal Large Language Models (MLLM) to the video domain, where the rich common sense priors in the language domain itself have not been fully exploited. 


The \textbf{key to extracting common sense} from language models is how to interpret visual activities in a way that natural language can interpret and distinguish. When we reason about subsequent activities, our predictions are heavily influenced by contextual information, such as what is happening (activity recognition), what is involved (object detection), and how these entities interact (human-object interaction). For example, in a \texttt{tidying} activity, a person holding the same object \texttt{broom} but with different human-object interactions, such as \texttt{picking up} and \texttt{holding}, is likely to lead to different activities: such as \texttt{putting it back in the closet} and \texttt{cleaning the floor}. Similarly, the same activities and interactions but with different objects may lead to different results. 
Therefore, in our framework, we first design a video context summary component to explicitly provide contextual summaries for video descriptions, which include candidate sets of objects, activities, and interactions between objects and activities. We then introduce common sense reasoning by using these three sets as prompt inputs to construct a description of the current situation (what is happening now), where irrelevant candidates can be naturally filtered through natural language generation. Taking a step further, we perform causal inference of subsequent actions (what will happen) by adding prompts ("The person then proceeds to") to the previously generated descriptions. 
We show in experiments that generating the current and subsequent descriptions plays a key role in yielding better performance. Our situation formulation simplifies existing work of spatial-temporal scene graph representations~\cite{AG} that require detailed and hierarchical segmentation of scene events. Instead of learning explicit graph connections between segments, we show that pre-trained large-scale language models (e.g. OPT~\cite{OPT}) can implicitly model correlations between objects, interactions and activities, and can naturally bring common sense priors through sentence generation.

Our final activity recognition head combines visual and textual cues to recognize video actions. In summary, our main \textbf{contributions} are:
\begin{itemize}
\item We introduce an activity recognition framework that involves common sense reasoning and casual inference through language-driven description generation of current and subsequent actions. 
\item We propose to explicitly model contextual summaries of activities, objects, and human-object interactions to extract common sense priors from language models.
\item We demonstrate the effectiveness of our proposed architecture on the challenging Action Genome and Charades datasets. 
\end{itemize}
\section{Related Work}
\label{sec::related}

\textbf{Video Action Recognition} aims to identify human actions from video sequences, and is widely used in tasks such as behavior analysis from surveillance cameras, augmented reality, and video editing.
Early established works are on trimmed videos containing a single action, demonstrating several widely used network architectures, such as two-stream networks~\cite{Simonyan2014TwoStream,Wang2019Pami}, 3D convolution networks~\cite{Tran15ICCV}, temporal convolutional networks (TCN), and convolution decomposition~\cite{I3D,Hussein2019CVPR}. 
With advances in understanding of trimmed single-action videos, the focus has begun to shift to multi-label activity recognition on longer and untrimmed videos of varying duration, occlusion, and lighting conditions, including architectures such as Slowfast~\cite{slowfast}, AssembleNet~\cite{AssembleNet} and Tokenlearner~\cite{Ryoo2021neurips}. 
Challenges remain, however, especially for in-the-wild monocular videos where cluttered sequences of daily activities and severe occlusion are observed. In this case, visual cues alone are not sufficient to recognize and differentiate actions.
Most recent video action recognition methods \cite{ju2022prompting,Qu_2024_CVPR, Maaz2023VideoChatGPT, Qu_2025_CVPR, azad2025hierarq} propose to adapt pre-trained MLLM to the video domain, showing excellent performance in the zero-shot setting.  However, the language model itself also contains rich common sense priors that are not fully utilized, which is the main research focus of this paper.

\textbf{Common sense reasoning} is a human-like ability to make presumptions about the type and essence of ordinary situations that humans encounter every day. 
When reading texts, humans make common sense inferences to frame their understanding of the presented narrative. 
In order to teach machines to acquire relevant and correct common sense in infinite situations, several common sense knowledge bases (KBs) such as ConceptNet \cite{ConceptNet}, Atomic \cite{ATOMIC}, WebChild \cite{webchild} and FB15k-237 have been established and has been widely used in past work to develop the common sense capability of machines. 
These KBs contain rich information on different aspects of common sense knowledge, e.g., lexical resources, time, space, quality, part-whole and aspirations/goals. 
Later, many works attempted to automatically construct common sense KBs. 
COMET \cite{COMET} is a common sense Transformer for automatically constructing knowledge graphs by training a language model on a seed set of knowledge tuples. 
Recently, large-scale language models trained on huge corpus have shown impressive capabilities including expressing common sense knowledge. 
In light of this, complex tasks (such as symbolic knowledge distillation~\cite{West2022SymbolicKD}) begin to leverage common sense knowledge in these large language models. 

\textbf{Prompt-based Learning.}
A prompt is a text string that contains some unfilled slots. 
Unlike traditional supervised learning, which trains a model to receive inputs and predict outputs, prompt-based learning is based on language models that directly model text probabilities. There are many prompt-based learning works that use these models to perform prediction tasks such as Vision-Language Models \cite{zhou2022cocoop,Lu2022CVPR,Radford2021LearningTV,Rao2022CVPR,XCLIP}, Vision Prompt \cite{Wang_2022_CVPR,jia2022vpt, Sun_2019_ICCV}, Pretrained Language Model \cite{NEURIPS2020Brown,Clark2020ELECTRA,tsimpoukelli2021}, Question Answering \cite{UNIFIEDQA} and Text Generation \cite{Schick2021Few}.
Prompt-based learning modifies the original input using a template prompt, and then uses the language model to probabilistically fill in the unfilled information, resulting in the final string, from which the final output can be derived.
This framework allows large language models (such as ChatGPT4o \cite{chatgpt4o}, Gemini \cite{google2024gemini2flash}, QWen \cite{bai2025qwen25vl} and OPT \cite{OPT}) pre-trained on massive amounts of raw text to be adapted to new scenarios with fewer or no labeled data. 
Recent advancements have extended prompt-based learning to video understanding tasks \cite{Jia24actionprompt, yang2025kronecker}.
By leveraging natural language prompts and task demonstrations as context, the model can well handle a wide range of tasks with only a few input examples, while not updating the parameters in the underlying model.

\begin{figure*}[!ht]
\centering
\begin{tabular}{c}
\includegraphics[width=\textwidth,keepaspectratio]{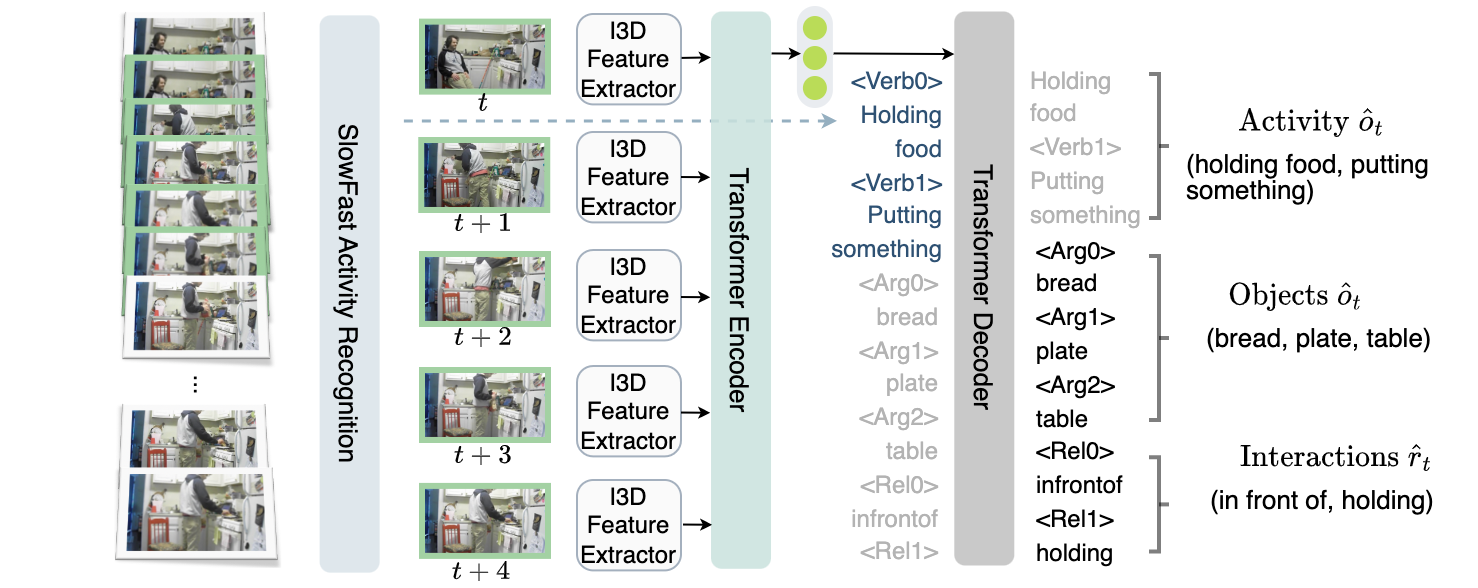}
\end{tabular}
\caption{
Overview of our Video Context Summary component $\mathcal{S}=\{\mathcal{S}_{int}, \mathcal{S}_{obj}, \mathcal{S}_{act}\}$. First, given a sequence of video frames $\{I_t\}^{T-1}_{t=0}$ sampled at a rate of $k$ fps, the SlowFast backbone identifies activities $\hat{v}$ over the entire video.
Then the frames $\{I_t\}_{t=0}^{T-1}$ are uniformly cropped into segments of length 5, denoted as $\{I_{t+i}\}_{i=0}^{4}$. Next, we take the segment $\{I_{t+i}\}_{i=0}^{4}$ as input and predict objects and human-object interactions for each frame. 
Specifically, we adopt a transformer-based sequence-to-sequence semantic role labeling model, where the I3D network is used to extract the visual features of each frame, and the transformer encoder encodes the extracted features and sends them to the transformer decoder. The transformer decoder is conditioned on the predicted activity verbs $\hat{v}$, takes the encoded features as input, and decodes the objects $\{\hat{o}_t\}^{T-1}_{t=0}$ and human-object interactions $\{\hat{r}_t\}^{T-1}_{t=0}$ for each frame.
}
\vspace{-5pt}
\label{fig:situation}
\end{figure*}

\begin{table*}[!ht]
\centering
\begin{tabular}
{|l|l|}
\hline
\rotatebox[origin=r]{90}{\parbox{6cm}{\centering Current Action Description Generation.}\quad\quad\quad } & \makecell{Generate a video caption based on the given human activities, objects and interactions. Example:
\\\\
Subject: person
\\Activities: Someone is undressing
\\Objects: clothes
Interactions: not looking at, in front of, holding
\\Video Caption: A person undresses in their closet, putting their clothes on the floor. They put a sweatsuit \\and some running shoes before inspecting themselves in the mirror.
\\\\
Subject: person
\\Activities: Closing a closet/cabinet
\\Objects: closet/cabinet, towel, blanket
\\Interactions: looking at, in front of, not contacting, holding
\\Video Caption: A person is seen taking a blanket out of a cabinet. They then open their closet put the blanket inside it.
\\\\
Subject: person
\\Activities: $<\text{Verb}_0>$,$<\text{Verb}_{1}>$, ..., $<\text{Verb}_{j}>$
\\Objects: $<\text{Arg}_0>$,$<\text{Arg}_{1}>$, ..., $<\text{Arg}_{m}>$ 
\\Interactions: $<\text{Rel}_0>$,$<\text{Rel}_{1}>$, ..., $<\text{Rel}_{l}>$ 
\\Video Caption: \color{Mahogany}{[$s^d$]}
}\\
\hline
\rotatebox[origin=r]{90}{\parbox{4cm}{\centering Subsequent Action Description Generation}} & 
\makecell{Predict the next activity using commonsense knowledge based on the given description. Examples:
\\\\
Description: The local meteorologist says it will rain.
\\The person then proceeds to take an umbrella.
\\\\
Description: The person saw a dirty plate and a red apple on the table.
\\The person then proceeds to clean up the kitchen.
\\\\
Description: <$s^d$>
\\The person then proceeds to \color{Mahogany}{[$s^c$]}
}\\
\hline
\end{tabular}
\caption{Prompt template used in the Commonsense Reasoning component (Sec. \ref{sec:commonsense}). In our current action description generation prompt template, we place a placeholder in angle brackets to be replaced by the video context summaries predicted in Sec. \ref{sec:situation}. In the prompt template for subsequent action inference, the angle brackets are the generated description of the current action. For both templates, the unfilled positions in square brackets are the target outputs that we want to fill with the language model. We also provide two additional examples for each prompt template, instructing the model to follow the same logic as the examples and generate sentences in the same format.}
\label{tab:prompt_template}
\end{table*}
\section{Method}
\label{sec::method}

Given a sequence of untrimmed video frames ${\bf{I}}=\{I_t\}^{T-1}_{t=0}$ sampled at a rate of $k$ fps, our task is to predict all the activities appeared in the input video $\hat{\bf{y}}$. 
Figure \ref{fig:overview} gives an overview of our framework. Our approach first generates contextual summaries of the entire video, including objects, activities, and interactions between objects and activities (Sec. \ref{sec:situation}). We then extract per-frame language descriptions based on the provided context, and inject additional prompts to infer subsequent activities through common sense reasoning (Sec.~\ref{sec:commonsense}). Finally, a multi-modal activity classifier combines visual and textual cues to recognize video actions (Sec. \ref{sec:multimodal}).

\subsection{Video Context Summarization}
\label{sec:situation}
We learn to generate contextual summaries of image descriptions, including: (1) main activities (e.g., holding some food), (2) objects (e.g., bread, plate) and (3) interactions between people and objects participating in the activity (e.g., looking at). In more detail, we predict context triples $\{(\hat{r}_t, \hat{o}_t, \hat{v}) | \hat{r}_t \subseteq R, \hat{o}_t \subseteq O, \hat{v} \subseteq V\}$ for each input frame $I_t$, where each item of the triple represents a candidate set of interaction, object and activity. For example, Figure \ref{fig:overview} shows the predicted situation as $(\hat{r}_t, \hat{o}_t, \hat{v})=$ ((\texttt{in front of}, \texttt{holding}, \texttt{looking at}), (\texttt{bread}, \texttt{plate}, \texttt{table}), (\texttt{Holding some food}, \texttt{Putting something on a table})).

Given that the situation triples are intrinsically related (e.g. objects and interactions are partially dependent on the activity over time), rather than predicting all three candidate sets simultaneously per frame, we propose to generate them in a hierarchical and sequential manner. Figure \ref{fig:situation} shows the detailed architecture of our design, as described below.

Our first stage, \textbf{activity recognition} module $\mathcal{S}_{act}$, predicts top-K candidate activities by looking at the entire video. We denote our multi-label prediction as a set of verbs $\hat{v}=\{\text{Verb}_0, \text{Verb}_1, ..., \text{Verb}_{K-1}\}$. The recognition module is pre-trained and frozen for downstream per-frame prediction tasks. Here, we choose to adopt one of the best performing backbone SlowFast~\cite{slowfast}. But note that our framework does not limit the use of any state-of-the-art recognizers over time.

Our second stage, \textbf{object and human-object interaction recognition} module $(\mathcal{S}_{obj}, \mathcal{S}_{int})$, predicts two candidate interactions $r_t$ and objects $o_t$ for each frame $I_t$ given the activity verbs of the video $\hat{v}$ as input. We adopt the sequence-to-sequence semantic role labeling model~\cite{vidsitu} as shown in Figure \ref{fig:situation}. For each video clip uniformly cropped from a video consisting of $l=5$ frames, we first extract the I3D features of the sequence, and then feed the features through a Transformer encoder to obtain a representation of the sequence.
Then, for each frame, the Transformer decoder takes the encoded representation of that frame and conditions on the active verbs $\hat{v}$ to generate a sequence of predicted objects and interactions.

\subsection{Common Sense Reasoning}
\label{sec:commonsense}
Our core component - the common sense reasoning component - aims to provide reasoning and causal inference for current and subsequent actions. We propose to leverage pre-trained large-scale language model to extract common sense priors through prompt-based learning.

\textbf{Current action description generation.}
We formulate our previously predicted per-frame video context summaries (from Sec. \ref{sec:situation}): human-object interactions $\hat{r}_t$, objects $\hat{o}_t$ and activities $\hat{v}$ as the prompt input to the description generation module $\mathcal{C}_{D}$, which generates the current frame scene description natural language sentences $s^d_t=\mathcal{C}_{D}(\hat{r}_t, \hat{o}_t, \hat{v})$. Specifically, $\mathcal{C}_{D}$ is guided to generate video descriptions according to the current video context description.  
Table~\ref{tab:prompt_template} shows the prompt template we designed to use as input. The template prompt has placeholders in angle brackets that will be replaced with the predicted verbs (activity), objects, and interactions $(\hat{r}_t, \hat{o}_t, \hat{v})$. The unfilled slots in the square brackets are our target output descriptions that we want to fill with the language model. There are also two examples provided in the prompt that guide the model to follow the same logic as the examples, producing sentences of the same format. 
As an example, in Figure \ref{fig:overview}, the generated current scene descriptions and reasoning is that $s^d_t=$\texttt{ A man is picking up a piece of bread...}.

\textbf{Subsequent action prediction.}
Given the descriptions of the current frame, we design $\mathcal{C}_{C}$ to perform causal inference using common sense knowledge implicitly learned by the language model, by predicting the next activity the person will do.
This can be achieved with additional prompt guidance as shown in Table \ref{tab:prompt_template}. The frame description statement $s^d_t$ is formulated as a prompt for input, preceded by the additional prompt statement "The person then proceeds to". The output is the subsequent action descriptions $s^c_t = \mathcal{C}_{C}(s^d_t)$ in natural language sentences. We also include two examples in the prompt template to guide predictions, which is the same as the previous current action description generation module. As an example, in Figure \ref{fig:overview}, our prediction of the next action is $s^c_t=$\texttt{ make a sandwich}. We show in experiments that this cascaded mode of current and subsequent action inference can show better performance gains.

\subsection{Multi-modal Activity Recognition}
\label{sec:multimodal}
Given the common sense prior implicitly included in the language descriptions predicted in Sec. \ref{sec:situation}, we can combine it with existing visual cues to predict the final video activity. 
We design a multi-modal activity recognition component $\mathcal{F}$ that first maps the input frame $I_t$, current frame descriptions $s^d_t$ and subsequent action reasoning $s^c_t$ to a common latent space. 
The learned mappings are and then fused as input to the activity classifier to predict the activity $\tilde{\bf{y}}_t = \mathcal{F}(I_t, s^d_t, s^c_t)$ at frame $t$. 

To model heterogeneous data from multiple modalities, we use CLIP encoders \cite{CLIP} to map the visual and textual inputs to the common latent space. 
The CLIP image encoder extracts the representation $\bf{u}$ of the input frame $I_t$. The CLIP text encoder extracts the representations $\bf{e}$ and $\bf{q}$ of the current frame description sentences $s^d_t$ and subsequent action sentences $s^c_t$, respectively. Then, the three representations are concatenated as input to the activity classifier to predict the final activity $\tilde{\bf{y}}_t = \mathcal{F}(I_t, s^d_t, s^c_t)$ at frame $t$. 
The predicted video activities $\hat{\bf{y}}$ are obtained by aggregating the predicted activities throughout the video $\{\tilde{\bf{y}}_t\}_{t=0}^{T-1}$. We also try to adopt other pre-trained multi-modal transformer model (e.g. FLAVA\cite{singh2022flava}), instead of using CLIP encoders, but observed slightly worse performance than CLIP.

Our final classifier is a three-layer MLP module with 512-512-N neurons at each layer, where N is the number of classes.  Since our text cues already contain temporal information about subsequent actions, we see no improvement in propagating embeddings from past frames (e.g. using LSTMs), thus conforming to the simple but effective MLP architecture.

\section{Experiments}
\label{sec:experiment}
We conduct experiments that compare with baselines and closely related work, showing quantitative and qualitative results as well as design-choice ablation studies.

\subsection{Experimental Setting}
\textbf{Baseline.}
To demonstrate that common sense reasoning can help improve video action recognition, we compare our full framework with the framework without common sense reasoning - the only activity recognition module with SlowFast~\cite{slowfast}.
We also compare to the closely related work using spatial-temporal scene graph SGFB~\cite{AG}.

\textbf{Datasets.}
Since our method is trained with annotated activities, objects and interactions, we evaluate the performance of our method on the Action Genome dataset \cite{AG} and the Charades dataset \cite{Charades}.
Charades \cite{Charades} consists of 9,848 annotated videos of 157 daily indoor activities with an average length of 30 seconds, where multiple overlapping activities may occur in each video.
Action Genome (AG) \cite{AG} is a large-scale video dataset built from Charades, providing human-object relationship labels in videos. It contains 10K videos with 234K frames annotated, with a total of 0.4 million objects of 35 object classes and 1.7 million instances of 25 visual relationship classes.

\textbf{Evaluation Metrics.} We use the metric mean Average Precision (mAP) to measure the performance.

\subsection{Implementation Details}
\textbf{Training.} We train our model on the Action Genome (AG) training set containing rich annotations of objects and interactions. In the video context summary component, we use SlowFast backbone pre-trained on Charades to predict top-5 video activity verbs $\hat{v} \subseteq V, |V|=157$ for each video. Given the groundtruth object-interaction tuple $(o_t, r_t)$ of the AG training set, the semantic role labeling model takes the activity verbs $\hat{v}$ as the input, and is trained to predict objects and interactions $\{(\hat{o}_t, \hat{r}_t)|\hat{o}_t \subseteq O, |O|=35, \hat{r}_t \subseteq R, |R|=25\}$ with cross-entropy loss for 30 epochs. We use the pre-trained large-scale OPT model \cite{OPT}, with 30B parameters to trade off quality and computational resources, as our language model in the common sense reasoning component of current and subsequent action descriptions: $\mathcal{C}_{D}$ and $\mathcal{C}_{C}$. 
The activity classifier $\mathcal{F}$ is trained on the AG training set, where each frame is annotated with ground truth activity labels.

\textbf{Inference.} We report the performance of the baselines and our approach on the validation set of the Action Genome and Charades datasets.
For Charades videos, we sample every 4 frames to predict frame activity $\tilde{\bf{y}}_t$. For all methods, we aggregate frame activities as our final video activity prediction output following the evaluation protocol.

\begin{figure}[!ht]
\centering
\begin{tabular}{c}
\includegraphics[width=0.45\textwidth,keepaspectratio]{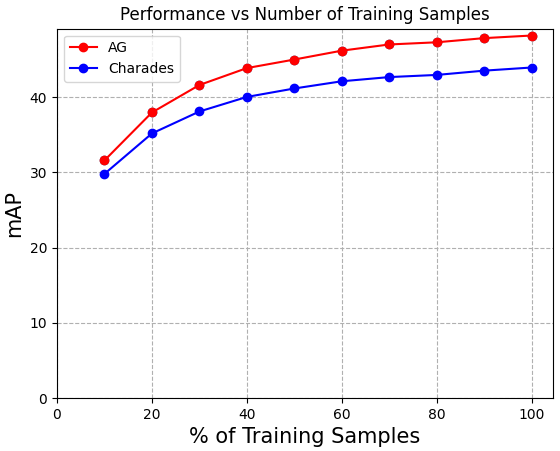}
\end{tabular}
\caption{
Generalization evaluation. Performance of our model on the test set when trained only using part of the training data. Our model can achieve 29.80\% mAP on the Charades dataset with only 10\% training examples.
}
\label{fig:fewshot}
\end{figure}

\begin{figure*}[ht!]
\centering
\renewcommand{\arraystretch}{2}
\rowcolors{2}{gray!25}{white}
\begin{tabular}{c p{4.2cm} p{4.2cm} p{4.2cm}}
\rowcolor{gray!50}
\parbox[c]{1.8cm}{\centering Method} & \parbox[c]{4.2cm}{\centering True Positives} & \parbox[c]{4.2cm}{\centering False Positives} & \parbox[c]{4.2cm}{\centering False Negatives}\\
\parbox[c]{1.8cm}{\centering SlowFast \cite{slowfast}} & \parbox[c]{4.2cm}{\small Wash a dish/dishes, Someone is cooking something} & \parbox[c]{4.2cm}{\small Washing something with a towel, Washing their hands} & \parbox[c]{4.2cm}{\small \textbf{Holding a dish, Putting a dish/es somewhere}, \textbf{Taking a dish/es from somewhere}} \\
\parbox[c]{1.8cm}{\centering CommonNet (ours)} & \parbox[c]{4.2cm}{\small Holding a dish, Putting a dish/es somewhere, Taking a dish/es from somewhere, Wash a dish/dishes, Someone is cooking something} & \parbox[c]{4.2cm}{\small \centering Washing their hands} & \parbox[c]{4.2cm}{\small \centering -} \\
\parbox[c]{1.8cm}{\centering Groundtruth} & \multicolumn{3}{c}{\parbox[c]{12.6cm}{\small Holding a dish, Putting a dish/es somewhere, Taking a dish/es from somewhere, Wash a dish/dishes, Someone is cooking something}} \\
\hline \hline
\rowcolor{gray!50}
 & \parbox[c]{4.2cm}{\centering Frame No. 163} & \parbox[c]{4.2cm}{\centering Frame No. 703} & \parbox[c]{4.2cm}{\centering Frame No. 919}\\
\parbox[c]{1.8cm}{\centering Selected Frames} & \ctab{\includegraphics[width=3cm,keepaspectratio]{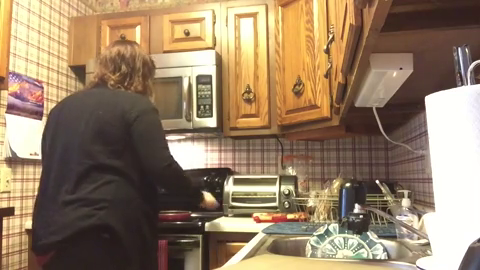}} &
\ctab{\includegraphics[width=3cm,keepaspectratio]{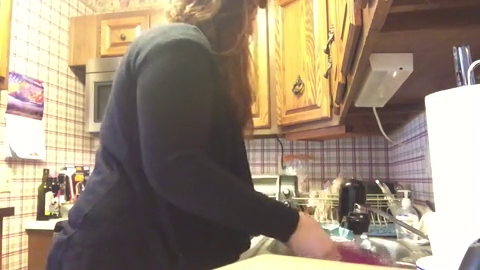}} & \ctab{\includegraphics[width=3cm,keepaspectratio]{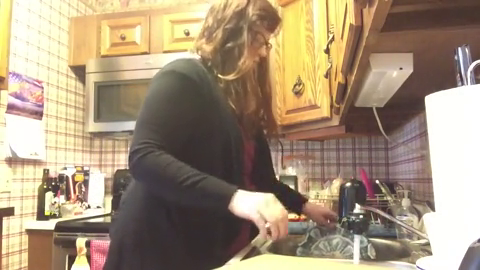}} \\
\hline \hline
\parbox[c]{1.8cm}{\centering Objects} & \parbox[c]{4.2cm}{\small food, dish} & \parbox[c]{4.2cm}{\small food, dish} & \parbox[c]{4.2cm}{\small food, dish} \\
\parbox[c]{1.8cm}{\centering Interactions} & \parbox[c]{4.2cm}{\small looking at, in front of, holding} & \parbox[c]{4.2cm}{\small looking at, in front of, holding} & \parbox[c]{4.2cm}{\small looking at, in front of, holding} \\
\parbox[c]{1.8cm}{\centering Description Generation} & \parbox[c]{4.2cm}{\small A person is seen \textbf{cooking something}, they then \textbf{take a dish} off the stove. They wash their hands. They then come into the kitchen where they are \textbf{holding a dish} in their left hand.} & \parbox[c]{4.2cm}{\small A person is seen \textbf{cooking something}, they also wash their hands before washing the dish/dishes of their hands.} & \parbox[c]{4.2cm}{\small A person is cooking something while they're also \textbf{washing dishes}. When the person's \textbf{cooking}, they're looking at the food. When they're \textbf{washing dishes}, they're \textbf{holding a dish} in their hand.}\\
\parbox[c]{1.8cm}{\centering Commonsense Reasoning} & \parbox[c]{4.2cm}{\small The person then proceeds to \textbf{put this dish} on the empty dining table.} & \parbox[c]{4.2cm}{\small The person then proceeds to \textbf{wash the dishes} for the pots that they cooked.} & \parbox[c]{4.2cm}{\small The person then proceeds to \textbf{put the clean dishes} in the dishwasher.}
\end{tabular}
\caption{Methods comparison on video W0QSB.mp4 in Action Genome dataset. Unlike the baseline method, which fails to detect complex activities such as \texttt{putting dishes somewhere}, our approach effectively constructs a coherent narrative timeline by leveraging descriptions of current actions ($\mathcal{C}_{D}$) and predicting subsequent activities ($\mathcal{C}_{C}$). From the first shot, $\mathcal{C}_{D}$ identifies a \texttt{cooking} activity involving \texttt{taking} and \texttt{holding} dishes, enabling $\mathcal{C}_{C}$ to accurately predict that the person will subsequently \texttt{put dishes down}. In the second shot, $\mathcal{C}_{D}$ continues to detect the ongoing \texttt{cooking} activity and notes the action \texttt{washing hands}, which prompts $\mathcal{C}_{C}$ to foresee the subsequent action of \texttt{washing dishes}. Finally, in the third shot, given the detected activity of \texttt{washing dishes} along with \texttt{holding dish}, $\mathcal{C}_{C}$ successfully predicts the action of \texttt{putting the clean dishes away} after washing.}
\vspace{-10pt}
\label{fig:qualitative1}
\end{figure*}

\begin{figure*}[ht!]
\centering
\renewcommand{\arraystretch}{2}
\rowcolors{2}{gray!25}{white}
\begin{tabular}{c p{4.2cm} p{4.2cm} p{4.2cm}}
\rowcolor{gray!50}
\parbox[c]{1.8cm}{\centering Method} & \parbox[c]{4.2cm}{\centering True Positives} & \parbox[c]{4.2cm}{\centering False Positives} & \parbox[c]{4.2cm}{\centering False Negatives}\\
\parbox[c]{1.8cm}{\centering SlowFast \cite{slowfast}} & \parbox[c]{4.2cm}{\small Closing a door, Opening a door, Walking through a doorway, Grasping onto a doorknob} & \parbox[c]{4.2cm}{\small Holding a phone/camera, Playing with a phone/camera, Closing a closet/cabinet, Opening a closet/cabinet, Someone is running somewhere} & \parbox[c]{4.2cm}{\small Holding a \textbf{book}, Holding a \textbf{box}} \\
\parbox[c]{1.8cm}{\centering CommonNet (ours)} & \parbox[c]{4.2cm}{\small Closing a door, Opening a door, Holding a book, Holding a box, Walking through a doorway, Grasping onto a doorknob} & \parbox[c]{4.2cm}{\small Taking a box from somewhere} & \parbox[c]{4.2cm}{\small \centering -} \\
\parbox[c]{1.8cm}{\centering Groundtruth} & \multicolumn{3}{c}{\parbox[c]{12.6cm}{\small Closing a door, Opening a door, Holding a book, Holding a box, Walking through a doorway, Grasping onto a doorknob}} \\
\hline \hline
\rowcolor{gray!50}
 & \parbox[c]{4.2cm}{\centering Frame No. 135} & \parbox[c]{4.2cm}{\centering Frame No. 219} & \parbox[c]{4.2cm}{\centering Frame No. 279}\\
\parbox[c]{1.8cm}{\centering Selected Frames} & \ctab{\includegraphics[width=3cm,keepaspectratio]{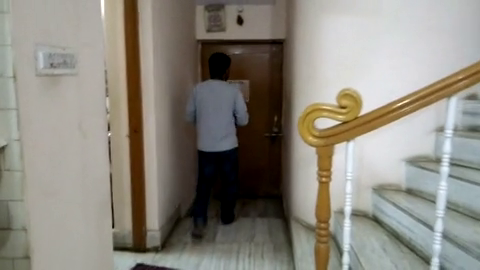}} &
\ctab{\includegraphics[width=3cm,keepaspectratio]{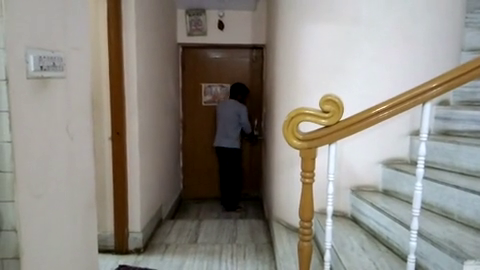}} & \ctab{\includegraphics[width=3cm,keepaspectratio]{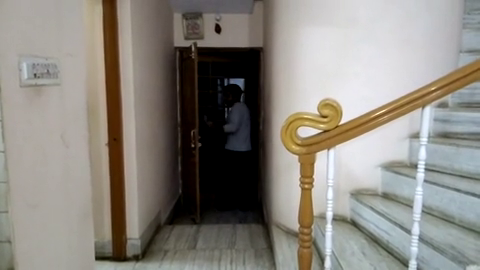}} \\
\hline \hline
\parbox[c]{1.8cm}{\centering Objects} & \parbox[c]{4.2cm}{\small doorknob, door} & \parbox[c]{4.2cm}{\small \textbf{book, box}, doorknob, door} & \parbox[c]{4.2cm}{\small doorway, door} \\
\parbox[c]{1.8cm}{\centering Interactions} & \parbox[c]{4.2cm}{\small in front of, holding} & \parbox[c]{4.2cm}{\small on the side of, holding, looking at, touching} & \parbox[c]{4.2cm}{\small in, not contacting} \\
\parbox[c]{1.8cm}{\centering Description Generation} & \parbox[c]{4.2cm}{\small A person is seen \texttt{through a doorway} being hugged and walked through a hallway.} & \parbox[c]{4.2cm}{\small A person opens their door. They then proceed to \texttt{place their hand on their doorknob}.} & \parbox[c]{4.2cm}{\small A person \texttt{walks across} the living room.}\\
\parbox[c]{1.8cm}{\centering Commonsense Reasoning} & \parbox[c]{4.2cm}{\small The person then proceeds to \texttt{open the white door}.} & \parbox[c]{4.2cm}{\small The person then proceeds to lock the door.} & \parbox[c]{4.2cm}{\small The person then proceeds to read a book.}
\end{tabular}
\caption{Methods comparison on video C0CMQ.mp4 in Action Genome dataset.}
\label{fig:qualitative2}
\end{figure*}

\subsection{Quantitative Results}

Table \ref{tab:comparison} shows the comparison between our baseline and our methods in Action Genome and Charades, respectively. Our method improves the mAP score by 18.99\% in the AG dataset and 13.13\% in the Charades dataset.

\setlength{\tabcolsep}{3pt}
\begin{table}[!ht]
\centering
\begin{tabular}
{l|l|l|l|l|l}
\toprule
\multirow{2}{*}{Method} & \multirow{2}{*}{Backbone}   & \multicolumn{4}{c}{mAP}   \\
& &  AG & $\Delta$ & Charades & $\Delta$ \\
\parbox{2cm}{SlowFast \cite{slowfast}} & \multirow{2}{*}{\parbox{1.5cm}{Res101-I3D-NL}} & - & \multirow{2}{*}{-} & 42.50 & \multirow{2}{*}{+1.80} \\
SGFB~\cite{AG} &  & - &   & 44.30 &  \\
\midrule
\midrule
\parbox{2cm}{SlowFast \cite{slowfast}}  & \multirow{2}{*}{\parbox{1.5cm}{Res50}} & 40.50 & \multirow{2}{*}{+7.69} & 38.84 & \multirow{2}{*}{+5.10}\\
Ours  &  & \bf{48.19} &   & \bf{43.94} & \\
\bottomrule
\end{tabular}
\caption{Comparison of activity recognition results using mAP on Action Genome and Charades datasets. Our method achieves better performance than that of our baseline SlowFast \cite{slowfast}. Moreover, comparing to the SGFB \cite{AG} that uses ResNet101-I3D-NL as backbone, we use the ResNet-50 as backbone but can achieve a larger improvement over the baseline.}
\label{tab:comparison}
\end{table}

We compare our approach to the closely related work of SGFB~\cite{AG}. We show that using language models can lead to relative higher performance gains compared to using spatial-temporal scene graphs. 
Also note that Ji et al. uses a more powerful ResNet-101 backbone for their SlowFast baseline, and while we choose ResNet-50 for code compatibility with language models, our proposed framework does not constrain us to use a better backbone.
Similar to SGFB, we evaluate the generalization property of our method as shown in Figure \ref{fig:fewshot}. We train our multi-modal activity classifier using only a subset of videos in the training set and compute mAP scores for the entire test set.
Our method can still achieve reasonable performance when trained with a very small amount of data.



\subsection{Qualitative Results}
Figure \ref{fig:qualitative1} and \ref{fig:qualitative2} compares our method to the baseline on two sample videos from the AG test set. We provide more examplex in the supplementary material. In Figure \ref{fig:qualitative1}, the video mainly involves two activities: cooking and washing dishes, and there are many sub-activities when moving the dishes during washing. 
The baseline method fails to detect sub-activities related to moving dishes, such as \texttt{putting a dish/es somewhere} and \texttt{taking a dish/es from somewhere}. While our method with common sense reasoning can detect all the activities without false positives and false negatives.
The main reason is that our language-driven description generation can cover most activities that benefit from the information provided by our object and interaction detectors, e.g., object \texttt{dish} and interaction \texttt{hold} -> \texttt{take a dish}. 
Also, common sense reasoning can help make reasonable predictions of the next step based on the descriptions, e.g., \texttt{take and hold a dish} -> \texttt{put the dish}. 
Similarly, in Figure \ref{fig:qualitative2}, the person is moving the box and book and walking through a doorway. 
Again, the baseline fails to detect these small movements such as \texttt{holding a book} and \texttt{holding a box}. Nevertheless, these false negatives of the baseline method can be detected by our method. Since our object detector has successfully detected the objects \texttt{book} and \texttt{box}, it is easier to detect the activities involving book and box.

\subsection{Ablation Study}
Table \ref{tab:ablation} shows ablation studies on different components. For the method using the image features only, we only use image embeddings as input to train our classifier.
For methods that use both image featuers and language descriptions, we concatenate image embeddings with the description embeddings and train a multi-modal classifier.
The method using all three features is the full method we describe in this paper. It can be seen that the method using all three components has the best performance compared to the method using only some of the components.

\begin{table}[!ht]
\centering
\renewcommand{\arraystretch}{1}
\begin{tabular}
{l|l|l}
\toprule
\multirow{2}{*}{Method} & \multicolumn{2}{c}{mAP} \\
& AG & Charades \\
\midrule
\parbox{4.5cm}{Image Features} & 22.14 & 21.92 \\
\midrule
\parbox{4.5cm}{Common Sense (subsequent)} & 23.57 & 23.61 \\
\midrule
\parbox{4.5cm}{Common Sense (current)} & 46.77 & 42.14 \\
\midrule
\parbox{4.5cm}{Image Features + Common Sense (current)} & 47.93 & 43.45 \\
\midrule
\parbox[c]{4.5cm}{Image Features + Common Sense (current + subsequent)} & \bf{48.19} & \bf{43.94} \\
\bottomrule
\end{tabular}
\caption{Ablation study of of different combinations of our components using mAP on Action Genome and Charades datasets. Here "current" in common sense means using features from the natural language description of the current action, and "subsequent" means the description of the next action.}
\label{tab:ablation}
\end{table}
\section{Limitations and Future Work}
Existing general-purpose common sense bases lack sufficient coverage and precision to meaningfully improve the recognition of domain-specific activities. 
For example, <\texttt{person}, \texttt{hold}, \texttt{folk}> under two different scenarios - restaurant and supermarket - can be recognized as "eat" and "checkout", respectively.
In the future, we plan to extend the current work to domain-specific common sense reasoning, introducing domain common sense priors into video understanding.

\section{Conclusions}
In this paper, we present an activity recognition framework that introduces common sense reasoning through language-driven description generation of current and subsequent actions. Experimental results on the challenging Action Genome and Charades datasets show the effectiveness of our approach.

{\small
\bibliographystyle{unsrt} 
\bibliography{cvpr}

\begin{thebibliography}{10}

\bibitem{AG}
Jingwei Ji, Ranjay Krishna, Li~Fei-Fei, and Juan~Carlos Niebles.
\newblock Action genome: Actions as compositions of spatio-temporal scene
  graphs.
\newblock In {\em ICCV}, pages 10236--10247, 2020.

\bibitem{OPT}
Susan Zhang, Stephen Roller, Naman Goyal, Mikel Artetxe, Moya Chen, Shuohui
  Chen, Christopher Dewan, Mona Diab, Xian Li, Xi~Victoria Lin, Todor Mihaylov,
  Myle Ott, Sam Shleifer, Kurt Shuster, Daniel Simig, Punit~Singh Koura, Anjali
  Sridhar, Tianlu Wang, and Luke Zettlemoyer.
\newblock {OPT}: Open pre-trained transformer language models.
\newblock {\em ArXiv}, abs/2205.01068, 2022.

\bibitem{RoBERTa}
Yinhan Liu, Myle Ott, Naman Goyal, Jingfei Du, Mandar Joshi, Danqi Chen, Omer
  Levy, Mike Lewis, Luke Zettlemoyer, and Veselin Stoyanov.
\newblock Roberta: {A} robustly optimized {BERT} pretraining approach.
\newblock {\em CoRR}, abs/1907.11692, 2019.

\bibitem{chatgpt4o}
OpenAI.
\newblock Chatgpt-4o, 2024.
\newblock Accessed: 2024-04-01.

\bibitem{google2024gemini2flash}
Google DeepMind.
\newblock Gemini 2.0 flash, 2024.
\newblock Accessed: 2025-04-01.

\bibitem{bai2025qwen25vl}
Shuai Bai, Keqin Chen, Xuejing Liu, Jialin Wang, Wenbin Ge, Sibo Song, Kai
  Dang, Peng Wang, Shijie Wang, Jun Tang, Humen Zhong, Yuanzhi Zhu, Mingkun
  Yang, Zhaohai Li, Jianqiang Wan, Pengfei Wang, Wei Ding, Zheren Fu, Yiheng
  Xu, Jiabo Ye, Xi~Zhang, Tianbao Xie, Zesen Cheng, Hang Zhang, Zhibo Yang,
  Haiyang Xu, and Junyang Lin.
\newblock Qwen2.5-vl technical report, 2025.
\newblock Accessed: 2025-04-01.

\bibitem{ju2022prompting}
Chen Ju, Tengda Han, Kunhao Zheng, Ya~Zhang, and Weidi Xie.
\newblock Prompting visual-language models for efficient video understanding.
\newblock In {\em European Conference on Computer Vision}, pages 105--124.
  Springer, 2022.

\bibitem{Qu_2024_CVPR}
Haoxuan Qu, Yujun Cai, and Jun Liu.
\newblock Llms are good action recognizers.
\newblock In {\em Proceedings of the IEEE/CVF Conference on Computer Vision and
  Pattern Recognition (CVPR)}, pages 18395--18406, June 2024.

\bibitem{Maaz2023VideoChatGPT}
Muhammad Maaz, Hanoona Rasheed, Salman Khan, and Fahad~Shahbaz Khan.
\newblock Video-chatgpt: Towards detailed video understanding via large vision
  and language models.
\newblock In {\em Proceedings of the 62nd Annual Meeting of the Association for
  Computational Linguistics (ACL 2024)}, 2024.

\bibitem{Qu_2025_CVPR}
Kai Hu, Feng Gao, Xiaohan Nie, Peng Zhou, Son Tran, Tal Neiman, Lingyun Wang,
  Mubarak Shah, Raffay Hamid, Bing Yin, and Trishul Chilimbi.
\newblock M-llm based video frame selection for efficient video understanding.
\newblock In {\em Proceedings of the IEEE/CVF Conference on Computer Vision and
  Pattern Recognition (CVPR)}, June 2025.

\bibitem{azad2025hierarq}
Shehreen Azad, Vibhav Vineet, and Yogesh~Singh Rawat.
\newblock Hierarq: Task-aware hierarchical q-former for enhanced video
  understanding.
\newblock {\em arXiv preprint arXiv:2503.08585}, 2025.

\bibitem{Simonyan2014TwoStream}
Karen Simonyan and Andrew Zisserman.
\newblock Two-stream convolutional networks for action recognition in videos.
\newblock In {\em Conference on Neural Information Processing Systems
  (Neurips)}, 2014.

\bibitem{Wang2019Pami}
Limin Wang, Yuanjun Xiong, Zhe Wang, Yu~Qiao, Dahua Lin, Xiaoou Tang, and Luc
  Van~Gool.
\newblock Temporal segment networks for action recognition in videos.
\newblock {\em IEEE Transactions on Pattern Analysis and Machine Intelligence},
  41(11):2740--2755, 2019.

\bibitem{Tran15ICCV}
Du~Tran, Lubomir Bourdev, Rob Fergus, Lorenzo Torresani, and Manohar Paluri.
\newblock Learning spatiotemporal features with 3d convolutional networks.
\newblock In {\em In ICCV}, 2015.

\bibitem{I3D}
Jo{\~{a}}o Carreira and Andrew Zisserman.
\newblock Quo vadis, action recognition? {A} new model and the kinetics
  dataset.
\newblock In {\em 2017 {IEEE} Conference on Computer Vision and Pattern
  Recognition, {CVPR} 2017, Honolulu, HI, USA, July 21-26, 2017}, 2017.

\bibitem{Hussein2019CVPR}
Noureldien Hussein, Efstratios Gavves, and Arnold~W.M. Smeulders.
\newblock Timeception for complex action recognition.
\newblock In {\em Proceedings of the IEEE/CVF Conference on Computer Vision and
  Pattern Recognition (CVPR)}, June 2019.

\bibitem{slowfast}
Christoph Feichtenhofer, Haoqi Fan, Jitendra Malik, and Kaiming He.
\newblock Slowfast networks for video recognition.
\newblock In {\em ICCV}, 2019.

\bibitem{AssembleNet}
Michael~S. Ryoo, A.~J. Piergiovanni, Juhana Kangaspunta, and Anelia Angelova.
\newblock Assemblenet++: Assembling modality representations via attention
  connections.
\newblock In {\em Computer Vision – ECCV 2020: 16th European Conference,
  Glasgow, UK, August 23–28, 2020, Proceedings, Part XX}, page 654–671,
  2020.

\bibitem{Ryoo2021neurips}
Michael Ryoo, AJ~Piergiovanni, Anurag Arnab, Mostafa Dehghani, and Anelia
  Angelova.
\newblock Tokenlearner: Adaptive space-time tokenization for videos.
\newblock In M.~Ranzato, A.~Beygelzimer, Y.~Dauphin, P.S. Liang, and J.~Wortman
  Vaughan, editors, {\em Advances in Neural Information Processing Systems},
  volume~34, pages 12786--12797, 2021.

\bibitem{ConceptNet}
Robyn Speer, Joshua Chin, and Catherine Havasi.
\newblock Conceptnet 5.5: An open multilingual graph of general knowledge.
\newblock {\em CoRR}, abs/1612.03975, 2016.

\bibitem{ATOMIC}
Maarten Sap, Ronan LeBras, Emily Allaway, Chandra Bhagavatula, Nicholas Lourie,
  Hannah Rashkin, Brendan Roof, Noah~A. Smith, and Yejin Choi.
\newblock {ATOMIC:} an atlas of machine commonsense for if-then reasoning.
\newblock {\em CoRR}, abs/1811.00146, 2018.

\bibitem{webchild}
Niket Tandon, Gerard de~Melo, and Gerhard Weikum.
\newblock {W}eb{C}hild 2.0 : Fine-grained commonsense knowledge distillation.
\newblock In {\em Proceedings of {ACL} 2017, System Demonstrations}, pages
  115--120, Vancouver, Canada, July 2017. Association for Computational
  Linguistics.

\bibitem{COMET}
Antoine Bosselut, Hannah Rashkin, Maarten Sap, Chaitanya Malaviya, Asli
  Celikyilmaz, and Yejin Choi.
\newblock {COMET}: Commonsense transformers for automatic knowledge graph
  construction.
\newblock In {\em Proceedings of the 57th Annual Meeting of the Association for
  Computational Linguistics}, pages 4762--4779, Florence, Italy, July 2019.
  Association for Computational Linguistics.

\bibitem{West2022SymbolicKD}
Peter West, Chandrasekhar Bhagavatula, Jack Hessel, Jena~D. Hwang, Liwei Jiang,
  Ronan~Le Bras, Ximing Lu, Sean Welleck, and Yejin Choi.
\newblock Symbolic knowledge distillation: from general language models to
  commonsense models.
\newblock In {\em NAACL}, 2022.

\bibitem{zhou2022cocoop}
Kaiyang Zhou, Jingkang Yang, Chen~Change Loy, and Ziwei Liu.
\newblock Conditional prompt learning for vision-language models.
\newblock In {\em IEEE/CVF Conference on Computer Vision and Pattern
  Recognition (CVPR)}, 2022.

\bibitem{Lu2022CVPR}
Yuning Lu, Jianzhuang Liu, Yonggang Zhang, Yajing Liu, and Xinmei Tian.
\newblock Prompt distribution learning.
\newblock In {\em Proceedings of the IEEE/CVF Conference on Computer Vision and
  Pattern Recognition (CVPR)}, pages 5206--5215, June 2022.

\bibitem{Radford2021LearningTV}
Alec Radford, Jong~Wook Kim, Chris Hallacy, Aditya Ramesh, Gabriel Goh,
  Sandhini Agarwal, Girish Sastry, Amanda Askell, Pamela Mishkin, Jack Clark,
  Gretchen Krueger, and Ilya Sutskever.
\newblock Learning transferable visual models from natural language
  supervision.
\newblock In {\em ICML}, 2021.

\bibitem{Rao2022CVPR}
Yongming Rao, Wenliang Zhao, Guangyi Chen, Yansong Tang, Zheng Zhu, Guan Huang,
  Jie Zhou, and Jiwen Lu.
\newblock Denseclip: Language-guided dense prediction with context-aware
  prompting.
\newblock In {\em Proceedings of the IEEE/CVF Conference on Computer Vision and
  Pattern Recognition (CVPR)}, pages 18082--18091, June 2022.

\bibitem{XCLIP}
Bolin Ni, Houwen Peng, Minghao Chen, Songyang Zhang, Gaofeng Meng, Jianlong Fu,
  Shiming Xiang, and Haibin Ling.
\newblock Expanding language-image pretrained models for general video
  recognition.
\newblock In {\em European Conference on Computer Vision (ECCV)}, 2022.

\bibitem{Wang_2022_CVPR}
Zifeng Wang, Zizhao Zhang, Chen-Yu Lee, Han Zhang, Ruoxi Sun, Xiaoqi Ren,
  Guolong Su, Vincent Perot, Jennifer Dy, and Tomas Pfister.
\newblock Learning to prompt for continual learning.
\newblock In {\em Proceedings of the IEEE/CVF Conference on Computer Vision and
  Pattern Recognition (CVPR)}, pages 139--149, June 2022.

\bibitem{jia2022vpt}
Menglin Jia, Luming Tang, Bor-Chun Chen, Claire Cardie, Serge Belongie, Bharath
  Hariharan, and Ser-Nam Lim.
\newblock Visual prompt tuning.
\newblock In {\em European Conference on Computer Vision (ECCV)}, 2022.

\bibitem{Sun_2019_ICCV}
Chen Sun, Austin Myers, Carl Vondrick, Kevin Murphy, and Cordelia Schmid.
\newblock Videobert: A joint model for video and language representation
  learning.
\newblock In {\em Proceedings of the IEEE/CVF International Conference on
  Computer Vision (ICCV)}, October 2019.

\bibitem{NEURIPS2020Brown}
Tom Brown, Benjamin Mann, Nick Ryder, Melanie Subbiah, Jared~D Kaplan, Prafulla
  Dhariwal, Arvind Neelakantan, Pranav Shyam, Girish Sastry, Amanda Askell,
  Sandhini Agarwal, Ariel Herbert-Voss, Gretchen Krueger, Tom Henighan, Rewon
  Child, Aditya Ramesh, Daniel Ziegler, Jeffrey Wu, Clemens Winter, Chris
  Hesse, Mark Chen, Eric Sigler, Mateusz Litwin, Scott Gray, Benjamin Chess,
  Jack Clark, Christopher Berner, Sam McCandlish, Alec Radford, Ilya Sutskever,
  and Dario Amodei.
\newblock Language models are few-shot learners.
\newblock In H.~Larochelle, M.~Ranzato, R.~Hadsell, M.F. Balcan, and H.~Lin,
  editors, {\em Advances in Neural Information Processing Systems}, volume~33,
  pages 1877--1901. Curran Associates, Inc., 2020.

\bibitem{Clark2020ELECTRA}
Kevin Clark, Minh-Thang Luong, Quoc~V. Le, and Christopher~D. Manning.
\newblock Electra: Pre-training text encoders as discriminators rather than
  generators.
\newblock In {\em International Conference on Learning Representations}, 2020.

\bibitem{tsimpoukelli2021}
Maria Tsimpoukelli, Jacob Menick, Serkan Cabi, SM~Eslami, Oriol Vinyals, and
  Felix Hill.
\newblock Multimodal few-shot learning with frozen language models.
\newblock {\em Proc. Neural Information Processing Systems}, 2021.

\bibitem{UNIFIEDQA}
Daniel Khashabi, Sewon Min, Tushar Khot, Ashish Sabharwal, Oyvind Tafjord,
  Peter Clark, and Hannaneh Hajishirzi.
\newblock {UNIFIEDQA}: Crossing format boundaries with a single {QA} system.
\newblock In {\em Findings of the Association for Computational Linguistics:
  EMNLP 2020}, Online, November 2020. Association for Computational
  Linguistics.

\bibitem{Schick2021Few}
Timo Schick and Hinrich Sch{\"u}tze.
\newblock Few-shot text generation with natural language instructions.
\newblock In {\em EMNLP}, 2021.

\bibitem{Jia24actionprompt}
Chengyou Jia, Minnan Luo, Xiaojun Chang, Zhuohang Dang, Mingfei Han, Mengmeng
  Wang, Guang Dai, Sizhe Dang, and Jingdong Wang.
\newblock Generating action-conditioned prompts for open-vocabulary video
  action recognition.
\newblock In {\em Proceedings of the 32nd ACM International Conference on
  Multimedia}, page 4640–4649, 2024.

\bibitem{yang2025kronecker}
Jingyi Yang, Zitong Yu, Xiuming Ni, Jia He, and Hui Li.
\newblock Kronecker mask and interpretive prompts are language-action video
  learners.
\newblock In {\em International Conference on Learning Representations}, 2025.

\bibitem{vidsitu}
Arka Sadhu, Tanmay Gupta, Mark Yatskar, Ram Nevatia, and Aniruddha Kembhavi.
\newblock Visual semantic role labeling for video understanding.
\newblock In {\em CVPR}, 2021.

\bibitem{CLIP}
Alec Radford, Jong~Wook Kim, Chris Hallacy, Aditya Ramesh, Gabriel Goh,
  Sandhini Agarwal, Girish Sastry, Amanda Askell, Pamela Mishkin, Jack Clark,
  Gretchen Krueger, and Ilya Sutskever.
\newblock Learning transferable visual models from natural language
  supervision.
\newblock {\em ArXiv}, abs/2103.00020, 2021.

\bibitem{singh2022flava}
Amanpreet Singh, Ronghang Hu, Vedanuj Goswami, Guillaume Couairon, Wojciech
  Galuba, Marcus Rohrbach, and Douwe Kiela.
\newblock Flava: A foundational language and vision alignment model.
\newblock In {\em Proceedings of the IEEE/CVF Conference on Computer Vision and
  Pattern Recognition}, pages 15638--15650, 2022.

\bibitem{Charades}
Gunnar~A. Sigurdsson, G{\"u}l Varol, X.~Wang, Ali Farhadi, Ivan Laptev, and
  Abhinav~Kumar Gupta.
\newblock Hollywood in homes: Crowdsourcing data collection for activity
  understanding.
\newblock {\em ArXiv}, abs/1604.01753, 2016.

\end{thebibliography}
}

\onecolumn
\begin{appendices}

\section{Ablation Study on Video Context Summary}
In this section, we show qualitative examples of current and next action description generation using different elements from video context summary in Table \ref{tab:supp_A1} and \ref{tab:supp_A2}. We show that using all the elements: activity, object and interaction, can generate a description that is closer to the ground-truth activities.

\begin{table*}[!ht]
\renewcommand{\arraystretch}{1.5}
\centering
\begin{tabular}
{|l|l|}
\hline
\parbox{2.1cm}{\centering \textbf{Video Context Summary}} & \textbf{Current and Next Action Description} \\
\hline
\multirow{2}{*}{\parbox{2cm}{\centering Activity}} & \parbox{0.85\textwidth}{$\mathcal{C}_{D}$: A person runs to the left, the person is not completely naked, they are wearing a long-sleeved sweatshirt and sweatpants. A person is undressing and then is fully naked, they are holding a dark blouse. In another scene a person is wearing clothes and \textit{\textbf{running}}, they then open a closet and put the clothes inside. Lastly, some undressing, while some is fully dressed and someone is holding clothes.
}\\ \cline{2-2}
& \parbox{0.85\textwidth}{$\mathcal{C}_{C}$: The person then proceeds to eat the lunch in the plate and takes a nap.} \\
\hline
\multirow{2}{*}{\parbox{2cm}{\centering Object}} & 
\parbox{0.85\textwidth}{$\mathcal{C}_{D}$: A person goes to a clothing rack and takes some shirts. After that, she takes a large shirt and puts it over her head by the shoulders, tying the sleeves. She moves her shoulders back and forth while looking at the mirror, as if she is wearing a mask.
}\\ \cline{2-2}
& \parbox{0.85\textwidth}{$\mathcal{C}_{C}$: The person then proceeds to turn and walk downstairs to the basement.} \\
\hline
\multirow{2}{*}{\parbox{2cm}{\centering Interaction}} & 
\parbox{0.85\textwidth}{$\mathcal{C}_{D}$: A person is seen on the side of their cabinet with their hand holding. The person doesn't look at anything but isn't holding anything.
}\\ \cline{2-2}
& $\mathcal{C}_{C}$: The person then proceeds to walk away. \\
\hline
\multirow{2}{*}{\parbox{2cm}{\centering Activity + Object}} & 
\parbox{0.85\textwidth}{$\mathcal{C}_{D}$: A person stands by a door. They open it and put their clothes on the dresser. They then put on some socks and a shirt. They walk away and they return, they are dressed.
}\\ \cline{2-2}
& \parbox{0.85\textwidth}{$\mathcal{C}_{C}$: The person then proceeds to clean out their closet before walking back upstairs.} \\
\hline
\multirow{2}{*}{\parbox{2cm}{\centering Activity + Interaction}} & 
\parbox{0.85\textwidth}{$\mathcal{C}_{D}$: A person is seen \textit{\textbf{running into the bedroom}}. While putting on their pants, they hear a noise. They look toward to find their brother. They give him a hug before putting on their shirt. Their other brother walks in with his jacket still on and his back to them. They give him a hug before he looks at them. They put a jacket on.
}\\ \cline{2-2}
& \parbox{0.85\textwidth}{$\mathcal{C}_{C}$: The person then proceeds to go to the living room.} \\
\hline
\multirow{2}{*}{\parbox{2cm}{\centering Object + Interaction}} & 
\parbox{0.85\textwidth}{$\mathcal{C}_{D}$: A person is standing at their bedroom door with several items in view (shoes, pants, shirts, underwear), placing his clothes on his bed.
}\\ \cline{2-2}
& \parbox{0.85\textwidth}{$\mathcal{C}_{C}$: The person then proceeds to walk from the bedroom to the bathroom.} \\
\hline
\multirow{2}{*}{\parbox{2cm}{\centering Activity + Object + Interaction}} & 
\parbox{0.85\textwidth}{$\mathcal{C}_{D}$: A person is seen carrying a sweater, putting it on. They \textit{\textbf{run past the dresser}} and \textit{\textbf{close the door}}. Another person sees the person looking at themselves in their closet and undress. The person is seen wearing a black shirt, putting clothes on a pile on the floor of their closet.
}\\ \cline{2-2}
& $\mathcal{C}_{C}$: The person then proceeds to take a towel and dry their hair. \\
\hline
\hline
\multicolumn{2}{|c|}{\parbox[c]{0.95\textwidth}{Ground-truth Activity: Grasping onto a doorknob, \textbf{Closing a door}, Someone is running somewhere}} \\
\hline
\multicolumn{2}{|c|}{\parbox[c]{0.95\textwidth}{Estimated Activity: Someone is running somewhere, Someone is undressing, Someone is dressing, Putting clothes somewhere, Holding some clothes}} \\
\hline
\multicolumn{2}{|c|}{\parbox[c]{0.95\textwidth}{Estimated Object: clothes, \textbf{door}}} \\
\hline
\multicolumn{2}{|c|}{\parbox[c]{0.95\textwidth}{Estimated Interaction: on the side of, \textbf{holding}, looking at, in front of, not contacting
}} \\
\hline
\end{tabular}
\vspace{-5pt}
\caption{Current and next action description generation using different elements from video context summary on the $165^{th}$ frame of video KRF68.mp4. The clauses that match the ground-truth activities are marked italic bold. We can see that using the activity, object and interaction as input can generate a sentence description more closer to the ground-truth activities. For example, the clause, "close the door", in the generated current action description is very close to the ground-truth activity "closing the door", which is very likely to benefit from the estimated object "door" and the estimated interaction "holding".}
\label{tab:supp_A1}
\end{table*}

\begin{table*}[!ht]
\renewcommand{\arraystretch}{1.5}
\centering
\begin{tabular}
{|l|l|}
\hline
\parbox{2.1cm}{\centering \textbf{Video Context Summary}} & \textbf{Current and Next Action Description} \\
\hline
\multirow{2}{*}{\parbox{2cm}{\centering Activity}} & \parbox{0.85\textwidth}{$\mathcal{C}_{D}$: A person \textit{\textbf{undresses}} in their bedroom, \textit{\textbf{putting their clothes in the closet}}. They then stand up and put on their shoes. They look out the window. The next day, the person is seen putting their clothes in the closet and then sitting on the couch.
}\\ \cline{2-2}
& \parbox{0.85\textwidth}{$\mathcal{C}_{C}$: The person then proceeds to undress again, but this time they are putting their clothes in the laundry basket.} \\
\hline
\multirow{2}{*}{\parbox{2cm}{\centering Object}} & 
\parbox{0.85\textwidth}{$\mathcal{C}_{D}$: A person is seen putting laptop on the couch and adjusting it position.
}\\ \cline{2-2}
& $\mathcal{C}_{C}$: The person then proceeds to play video game. \\
\hline
\multirow{2}{*}{\parbox{2cm}{\centering Interaction}} & 
\parbox{0.85\textwidth}{$\mathcal{C}_{D}$: a person is not looking at anything at the bottom of a bath tub, and is sitting on a plastic bag. They then turn on the sink and wash their hands. They are in front of a wall with a mirror over top of it.
}\\ \cline{2-2}
& $\mathcal{C}_{C}$: The person then proceeds to then dry themselves off with a face towel. \\
\hline
\multirow{2}{*}{\parbox{2cm}{\centering Activity + Object}} & 
\parbox{0.85\textwidth}{$\mathcal{C}_{D}$: A person \textit{\textbf{undresses}} in their bedroom. They're seen getting ready to go outside before sitting on their couch and using their laptop.
}\\ \cline{2-2}
& \parbox{0.85\textwidth}{$\mathcal{C}_{C}$: The person then proceeds to sit online and work.} \\
\hline
\multirow{2}{*}{\parbox{2cm}{\centering Activity + Interaction}} & 
\parbox{0.85\textwidth}{$\mathcal{C}_{D}$: A person is seen \textit{\textbf{undressing}} when they go out in their backyard. They continue to change as they open a window that overlooks the sea. They continue to look out of the window while sitting on a couch. They then decide to \textit{\textbf{put their clothes somewhere}}. They take off their top and look underneath their couch to see where their clothes went.
}\\ \cline{2-2}
& \parbox{0.85\textwidth}{$\mathcal{C}_{C}$: The person then proceeds to get out some sort of device and starts talking to the people gathered in front of him.} \\
\hline
\multirow{2}{*}{\parbox{2cm}{\centering Object + Interaction}} & 
\parbox{0.85\textwidth}{$\mathcal{C}_{D}$: A person is looking at their laptop. They lean back on the sofa.
}\\ \cline{2-2}
& $\mathcal{C}_{C}$: The person then proceeds to browse Facebook. \\
\hline
\multirow{2}{*}{\parbox{2cm}{\centering Activity + Object + Interaction}} & 
\parbox{0.85\textwidth}{$\mathcal{C}_{D}$: A person \textit{\textbf{takes their clothes off}}. They \textit{\textbf{put them into a laundry basket}} before heading to the front door and \textit{\textbf{putting out a bunch of lights}} on the way.
}\\ \cline{2-2}
& \parbox{0.85\textwidth}{$\mathcal{C}_{C}$: The person then proceeds to open the front door and closes it as well.} \\
\hline
\hline
\multicolumn{2}{|c|}{\parbox[c]{0.95\textwidth}{Ground-truth Activity: \textbf{Turning off a light}, \textbf{Someone is undressing}, \textbf{Throwing clothes somewhere}}} \\
\hline
\multicolumn{2}{|c|}{\parbox[c]{0.95\textwidth}{Estimated Activity: Someone is undressing, Watching/Looking outside of a window, Sitting on sofa/couch, Someone is dressing, Putting clothes somewhere}} \\
\hline
\multicolumn{2}{|c|}{\parbox[c]{0.95\textwidth}{Estimated Object: \textbf{light}, \textbf{clothes}}} \\
\hline
\multicolumn{2}{|c|}{\parbox[c]{0.95\textwidth}{Estimated Interaction: on the side of, not contacting, looking at, in front of, holding}} \\
\hline
\end{tabular}
\caption{Current and next action description generation using different elements from video context summary on the $611^{th}$ frame of video Y79PC.mp4. The clauses that match the ground-truth activities are marked italic bold. The sentence that are generated using the activity, object and interaction contains clauses matched to all of the ground-truth activities. For example, the generated clauses, "putting out a bunch of lights", "takes their clothes off" and "put them into a laundry basket", are matched to the ground-truth activities "Turning off a light", "Someone is undressing" and "Throwing clothes somewhere", respectively. This is due to that both of the objects, "light" and "clothes", are successfully detected.}
\label{tab:supp_A2}
\end{table*}

\section{More Qualitative Results}

\begin{figure*}[ht!]
\centering
\renewcommand{\arraystretch}{2}
\rowcolors{2}{gray!25}{white}
\begin{tabular}{c p{4.2cm} p{4.2cm} p{4.2cm}}
\rowcolor{gray!50}
\parbox[c]{1.8cm}{\centering Method} & \parbox[c]{4.2cm}{\centering True Positives} & \parbox[c]{4.2cm}{\centering False Positives} & \parbox[c]{4.2cm}{\centering False Negatives}\\
\parbox[c]{1.8cm}{\centering SlowFast \cite{slowfast}} & \parbox[c]{4.2cm}{\small Holding some food, Taking food from somewhere, Holding a dish, Putting a dish/es somewhere, Taking a dish/es from somewhere, Someone is cooking something, Someone is eating something} & \parbox[c]{4.2cm}{\small Holding a cup/glass/bottle of something} & \parbox[c]{4.2cm}{\small \textbf{Putting something on a table}, \textbf{Putting some food somewhere}} \\
\parbox[c]{1.8cm}{\centering CommonNet (ours)} & \parbox[c]{4.2cm}{\small Putting something on a table, Holding some food, Putting some food somewhere, Taking food from somewhere, Holding a dish, Putting a dish/es somewhere, Taking a dish/es from somewhere, Someone is cooking something, Someone is eating something} & \parbox[c]{4.2cm}{\small \centering -} & \parbox[c]{4.2cm}{\small \centering -} \\
\parbox[c]{1.8cm}{\centering Groundtruth} & \multicolumn{3}{c}{\parbox[c]{12.6cm}{\small Putting something on a table, Holding some food, Putting some food somewhere, Taking food from somewhere, Holding a dish, Putting a dish/es somewhere, Taking a dish/es from somewhere, Someone is cooking something, Someone is eating something}} \\
\hline \hline
\rowcolor{gray!50}
 & \parbox[c]{4.2cm}{\centering Frame No. 59} & \parbox[c]{4.2cm}{\centering Frame No. 199} & \parbox[c]{4.2cm}{\centering Frame No. 431}\\
\parbox[c]{1.8cm}{\centering Selected Frames} & \ctab{\includegraphics[width=3cm,keepaspectratio]{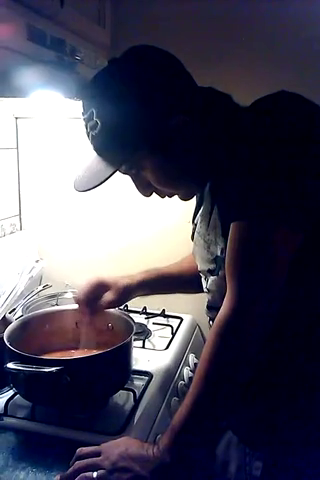}} &
\ctab{\includegraphics[width=3cm,keepaspectratio]{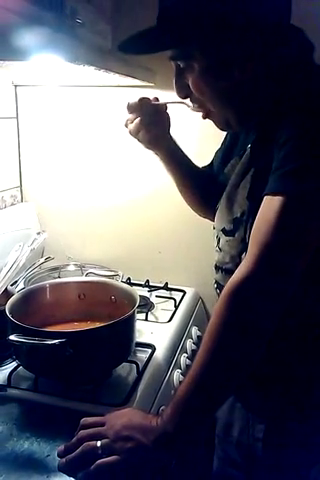}} & \ctab{\includegraphics[width=3cm,keepaspectratio]{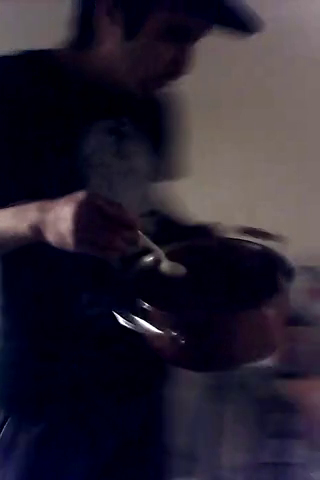}} \\
\hline \hline
\parbox[c]{1.8cm}{\centering Objects} & \parbox[c]{4.2cm}{\small food, dish, \textbf{table}} & \parbox[c]{4.2cm}{\small food} & \parbox[c]{4.2cm}{\small food, dish} \\
\parbox[c]{1.8cm}{\centering Interactions} & \parbox[c]{4.2cm}{\small looking at, in front of, \textbf{holding}} & \parbox[c]{4.2cm}{\small looking at, in front of} & \parbox[c]{4.2cm}{\small looking at, in front of, holding} \\
\parbox[c]{1.8cm}{\centering Description Generation} & \parbox[c]{4.2cm}{\small A person is seen \textbf{holding a plate} and looking at them. They are then shown \textbf{eating} something from the plate. A chef prepares food and \textbf{places it on the table}. A person then is shown \textbf{taking some food from the plate}.} & \parbox[c]{4.2cm}{\small A person is seen \textbf{holding a plate with some food}.} & \parbox[c]{4.2cm}{\small A person is seen \textbf{putting some food on a plate}. They then close-up of that food. They open-up the top to the drawer, then take a plate out of it, also the person sees the food, \textbf{holding} onto it.}\\
\parbox[c]{1.8cm}{\centering Commonsense Reasoning} & \parbox[c]{4.2cm}{\small The person then proceeds to \textbf{eat the food}.} & \parbox[c]{4.2cm}{\small The person then proceeds to \textbf{eat} without eating out of a bowl, instead using their hands.} & \parbox[c]{4.2cm}{\small The person then proceeds to \textbf{take the platter} in one hand and the bag in the other hand and \textbf{puts the platter on the table}.}
\end{tabular}
\caption{Methods comparison on video 2B577.mp4.}
\label{fig:qualitative3}
\end{figure*}

Figure \ref{fig:qualitative3} shows an additional sample of our results compared to the baseline method. We can see that the person is cooking, taking the food from the kitchen to the dinning room and then put the food on the table. The baseline fails to detect "put something on a table" and "put some food somewhere". On the contrary, our method successfully detected the objects "food", "dish" and "table", and the interactions "holding" and "in front of", which gives a scene that a person is holding food and plate in front of a table. Then the commonsense reasoning model successfully predicts that the person is "putting something on the table".

\section{More Implementation Details}
We use the OPT-30B version as our action description generator. We set the maximum length of the generated tokens to be 260 and 200 for the current and next action description generators, respectively.

\end{appendices}

\end{document}